\documentclass{article}
\usepackage{iclr2026_conference,times}
\usepackage{hyperref}
\usepackage{url}
\usepackage{amsmath,amssymb}
\usepackage{booktabs}
\usepackage{graphicx}
\usepackage{subcaption}
\usepackage{algorithm}
\usepackage[noend]{algpseudocode}
\usepackage{mathtools}
\usepackage{microtype}
\usepackage{xcolor}
\usepackage{comment}
\usepackage{multirow}
\usepackage{tabularx}
\usepackage[margin=1in]{geometry}
\usepackage{array}
\usepackage{graphicx} 
\usepackage{caption}
\usepackage{adjustbox}

\title{StateFlow: Dual-State Recurrent Modeling for Long-Horizon time series Forecasting}

\author{
Anonymous Authors\\
Anonymous Affiliation\\
\texttt{anonymous@anonymous.edu}
}

\author{
\textbf{Haroon Gharwi\textsuperscript{1*} \quad Yue Dai\textsuperscript{2*} \quad Kai Shu\textsuperscript{3+}}\\[3pt]
\textsuperscript{*}Department of Computer Science, Illinois Institute of Technology, Chicago, IL, USA \\
\textsuperscript{+}Department of Computer Science, Emory University, Atlanta, GA, USA\\[3pt]
\textsuperscript{1}\texttt{hgharwi@hawk.illinoistech.edu} \quad
\textsuperscript{2}\texttt{ydai21@illinoistech.edu} \quad
\textsuperscript{3}\texttt{kai.shu@emory.edu}
}

\iclrfinalcopy  

\newcommand{\ignore}[1]{}

\newcommand{\keywords}[1]{%
  \par\vspace{0.5\baselineskip}%
  \noindent\textbf{Keywords—} #1\par}
  
\makeatletter
\renewcommand{\@thanks}{}
\renewcommand{\@oddhead}{}
\renewcommand{\@evenhead}{}
\makeatother

\usepackage{fancyhdr}
\fancypagestyle{varnnstyle}{
  \fancyhf{} 
  \rfoot{\small Preprint. © 2026} 
  \cfoot{\thepage}               
  \renewcommand{\headrulewidth}{0pt}
  \renewcommand{\footrulewidth}{0pt}
}
\begin{document}
\maketitle

\begin{abstract}
Long-horizon multivariate time series forecasting (LTSF) remains challenging due to non-stationarity, regime shifts, and error accumulation. 
The Variability-Aware Recursive Neural Network (VARNN) was recently introduced to track such variability by maintaining a residual-memory state driven by one-step prediction errors. 
However, its original formulation is limited to one-step sequence regression 
and does not directly support multi-step forecasting.
In this work,\ignore{To address this gap} we extend VARNN to long-horizon forecasting and introduce StateFlow, a recurrent forecasting framework that uses VARNN as a dual-state recurrent backbone to capture two complementary signals from the lookback sequence: a hidden-state trajectory representing primary temporal dynamics, including trend, seasonality, level changes, and recurring patterns, and a residual-memory trajectory representing structured local prediction deviations, driven from a nonlinear recurrent transformation of errors between one-step base predictions and observed values. A chunk-based decoder separately summarizes these trajectories and maps them to the future horizon for direct multi-step forecasting. We further employ a two-stage optimization strategy that first trains the VARNN encoder through a one-step base prediction objective to optimize\ignore {residual-aware recurrent} the internal representations over the lookback sequence, and then trains a horizon-specific decoder for direct multi-step forecasting. Experiments on standard LTSF benchmarks show that StateFlow achieves competitive performance against strong linear, recurrent, convolutional, and Transformer-based baselines while preserving linear recurrent encoding and a compact model design.

\end{abstract}


\section{Introduction}
Long-horizon multivariate time series forecasting (LTSF) is a fundamental problem in applications such as energy demand prediction, traffic flow analysis, financial modeling, and environmental monitoring~\citep{hyndman2021fpp3, brockwell2002introTS, twentyfive_ts}. LTSF requires modeling extended temporal dependencies while remaining robust to non-stationarity, distribution shift, and evolving seasonal structure~\citep{lim2021deeplearning, baidya2024nonstationarity}. As the prediction horizon increases, modeling errors tend to accumulate, making stable long-term extrapolation particularly challenging~\citep{lim2019tft, oreshkin2019nbeats, benidis2023survey}. \ignore{These challenges have motivated extensive investigation into deep learning approaches specifically designed for long-horizon forecasting settings~\citep{zeng2023dlinear}.}

Recent progress in LTSF has been largely driven by Transformer-based architectures that leverage self-attention to capture long-range dependencies. Patch-based tokenization strategies and channel-wise attention mechanisms have demonstrated strong empirical performance on standard benchmarks~\citep{nie2023patchtst, liu2024itransformer}. However, attention-based models generally exhibit quadratic complexity with respect to input length and often rely on global interaction patterns that may introduce computational overhead when long lookback windows are required. 

Recurrent architectures, in contrast, scale linearly with sequence length and provide a natural causal structure for temporal modeling. Classical recurrent models such as RNNs, LSTMs, and GRUs~\citep{elman1990rnn, hochreiter1997lstm, cho2014gru} have been widely studied for sequence regression. Nevertheless, their performance in long-horizon forecasting settings has often been observed to lag behind modern Transformer-based methods~\citep{zhou2021informer, zeng2023dlinear}. One potential limitation is that conventional recurrence primarily captures content evolution through a single hidden state and does not explicitly model systematic deviation\ignore{over the lookback observations}. These local deviations may reflect drift, bias, or other non-stationary changes in the observed sequence.

Recent work introduced the Variability-Aware Recursive Neural Network (VARNN) for one-step sequence regression~\cite{gharwi2025varnn}. VARNN explicitly uses one-step prediction residuals to update a dedicated nonlinear residual-memory state, allowing structured local prediction deviations to be modeled separately from primary temporal dynamics. However, its original formulation is limited to one-step sequence regression, and its applicability to long-horizon time series forecasting (LTSF) remains unexplored. 

To address this gap, we extend VARNN to the LTSF setting and introduce the StateFlow framework. StateFlow uses VARNN as a dual-state recurrent backbone to capture two complementary signals from the lookback sequence: a hidden-state trajectory representing primary temporal dynamics, including trend, seasonality, level changes, and recurring patterns, and a residual-memory trajectory representing structured local prediction deviations, driven from recurrent nonlinear transformation of errors between one-step base predictions and observed lookback values. A chunk-based forecasting decoder separately summarizes these trajectories before mapping them to the future horizon. This design preserves the distinction between temporal and deviation dynamics while reducing the parameter growth associated with directly flattening the full encoder trajectories.

\begin{figure}[t]
    \centering
    \includegraphics[width=0.60\linewidth]{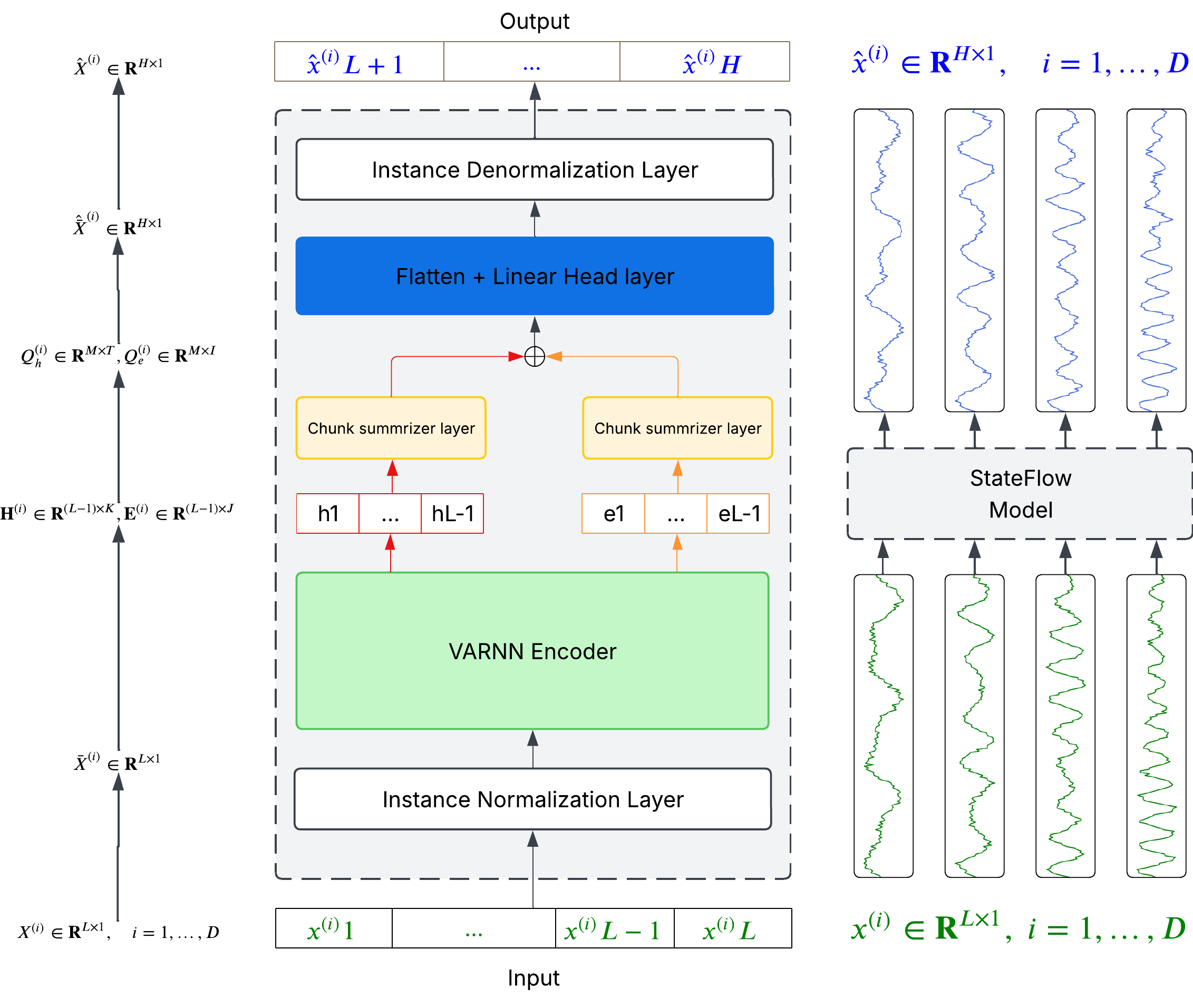}
    \caption{Architecture of the proposed StateFlow framework with VARNN as the recurrent backbone. VARNN encodes the lookback sequence into hidden-state and residual-memory trajectories, which are then separately summarized by a chunk-based decoder and projected to the future horizon for direct multi-step forecasting.}
    \label{fig:VARNN}
\end{figure}
We further employ a two-stage optimization strategy to bridge one-step residual learning and long-horizon forecasting. In the first stage, the VARNN encoder is trained using a one-step prediction objective over the lookback window, giving the residual-memory state an explicit learning signal. In the second stage, the pretrained encoder is frozen and a horizon-specific decoder is trained for direct multi-step forecasting. This separation stabilizes optimization and allows the same pretrained VARNN encoder to be reused across multiple prediction horizons for a fixed dataset and lookback length, while a separate horizon-specific decoder is trained for each horizon.

The proposed StateFlow model preserves linear computational complexity with respect to input length through its recurrent encoder. Our empirical results on standard LTSF benchmarks show that StateFlow achieves competitive performance against strong linear, convolutional, recurrent, and Transformer-based baselines, while avoiding quadratic temporal self-attention. These results suggest that incorporating deviation dynamics into recurrence provides an efficient alternative to attention-based LTSF models for long-horizon forecasting.
Our contributions are summarized as follows:
\begin{itemize}
\item We investigate whether recurrence can remain competitive for LTSF when deviation dynamics are explicitly incorporated into the state transition. 
\item We extend VARNN from one-step sequence regression to long-horizon time-series forecasting by using its hidden and residual-memory trajectories as dual-state forecasting representations.
\item We introduce a StateFlow framework, a recurrent trajectory forecasting framework with a chunk-based horizon decoder that separately summarizes hidden dynamics and residual-memory dynamics before direct multi-step prediction.
\item We employ a two-stage optimization strategy that first learns residual-aware recurrent representations through one-step prediction and then trains horizon-specific decoders for long-horizon forecasting.
\item We provide systematic experiments and ablations showing that residual-memory recurrence is a competitive and parameter-efficient inductive bias for long-horizon forecasting.
\end{itemize}

\section{Related Work}

\subsection{Statistical and Linear Models}
Classical forecasting methods, including ARIMA and exponential smoothing, provide principled frameworks for modeling temporal structure under linearity and stationarity assumptions~\citep{box2015time,hyndman2021fpp3}. Although effective in many settings, these methods may be limited in capturing complex nonlinear and non-stationary dynamics. Recent work has revisited lightweight linear architectures for long-horizon time-series forecasting.  LTSF-Linear and DLinear demonstrate that carefully structured linear mappings, combined with suitable decomposition or normalization strategies, can outperform several more complex Transformer-based models on standard benchmarks~\citep{zeng2023dlinear}. RLinear further highlights the importance of reversible normalization and channel-independent modeling in the performance of linear forecasting methods~\citep{li2023rlinear}. These findings suggest that carefully chosen temporal inductive biases can be as important as architectural complexity in long-horizon forecasting.

\subsection{Transformer-Based Models}

Transformer architectures~\citep{vaswani2017attention} have been widely adopted for time series forecasting due to their ability to capture long-range dependencies through self-attention mechanisms. Informer~\citep{zhou2021informer} improves efficiency using sparse attention, while Autoformer and FEDformer incorporate decomposition and frequency-domain modeling~\citep{wu2021autoformer,zhou2022fedformer}, PatchTST introduces patch-based tokenization with channel-independent forecasting strategies~\citep{nie2023patchtst}, and iTransformer applies attention across variables by inverting the temporal and variate dimensions~\citep{liu2024itransformer}. These approaches have achieved strong empirical performance on long-horizon benchmarks such as ETT, Weather, and Electricity datasets. Despite their effectiveness, attention-based models typically retain quadratic complexity in sequence length or require additional approximations to reduce this cost. Moreover, their reliance on global interaction patterns can introduce additional overhead when long lookback windows are used. This motivates exploring alternative sequential inductive biases that preserve efficiency while remaining competitive.

\subsection{Recurrent Neural Networks}

Recurrent neural networks, including vanilla RNNs~\cite{elman1990rnn}, long short-term memory networks (LSTMs)~\cite{hochreiter1997lstm}, and gated recurrent units (GRUs)~\cite{cho2014gru} model nonlinear autoregressive temporal dynamics through hidden-state transitions. They provide a natural causal structure for sequence modeling and scale linearly with input length. However, these conventional recurrent models typically compress temporal information into a single hidden state that integrates past information implicitly, which may limit their ability to explicitly track systematic prediction deviations under non-stationarity.

Recently, the Variability-Aware Recursive Neural Network (VARNN)~\cite{gharwi2025varnn} was introduced to explicitly model such deviations in one-step sequence regression with covariate inputs. Unlike conventional recurrent models that encode the covariate sequence only through latent states, VARNN additionally leverages the historical ground-truth of observed sequence and computes the residual between each one-step base prediction and its corresponding observed value, then transforms these residuals via a nonlinear recurrent update into a dedicated residual-memory trajectory that conditions the subsequent step. Prior results suggest that residuals can exhibit structured nonlinear dependencies that provide useful information for modeling temporal deviations. However, VARNN was originally formulated for one-step regression, and its effectiveness for long-horizon time-series forecasting has not been systematically studied. In this work, we extend VARNN to the LTSF setting by using its hidden-state and residual-memory trajectories as forecasting representations and incorporating a chunk-based horizon decoder with two-stage optimization for direct multi-step forecasting.

\section{Methodology}
\label{sec:methodology}
\subsection{Problem Formulation}

Let $\mathbf{X}_{1:L} = [\mathbf{x}_1, \mathbf{x}_2, \ldots, \mathbf{x}_L] \in \mathbb{R}^{L \times D}$ denote a multivariate input sequence with lookback length $L$ and $D$ variables, where $\mathbf{x}_t \in \mathbb{R}^{D}$ is the observation at time step $t$. The objective of long-horizon forecasting is to predict the future sequence 
$
    \mathbf{Y}_{L+1:L+H} = [\mathbf{x}_{L+1}, \mathbf{x}_{L+2}, \ldots, \mathbf{x}_{L+H}]
    \in \mathbb{R}^{H \times D},
$
where $H$ is the forecasting horizon. 

\subsection{Model Architecture}
The proposed StateFlow architecture consists of three main components: an instance normalization layer, a dual-state recurrent encoder, and a horizon forecasting decoder. To improve scalability across multivariate datasets, the model follows a channel-independent processing strategy. Each variable is processed as an independent univariate sequence while sharing the same model parameters across variables. We split the multivariate time series input $\mathbf{X} \in \mathbb{R}^{ L \times D}$, to $D$ univariate series $\mathbf{X^{(i)} }_{1:L} = [\mathbf{x^{(i)} }_1, \mathbf{x^{(i)} }_2, \ldots, \mathbf{x^{(i)} }_L] \in \mathbb{R}^{1 \times L}$, where i = 1,\ldots,D.  Accordingly, the model will predict  $
    \mathbf{\hat{Y}^{(i)} }_{L+1:L+H} = [\mathbf{x^{(i)} }_{L+1}, \mathbf{x^{(i)} }_{L+2}, \ldots, \mathbf{x^{(i)} }_{L+H}]
    \in \mathbb{R}^{1\times H}.
    $ 
    After forecasting, the output is reshaped back to
    $\hat{\mathbf{Y}} \in \mathbb{R}^{H \times D}$.

\subsubsection{Instance Normalization Layer}
To reduce window-level distribution shifts between training and test in non-stationary time series, we apply instance normalization to each input sequence $\mathbf{X}_{1:L}$ before encoder processing \cite{ulyanov2016instance, kim2022revin}. For each variable, the mean and standard deviation are computed over the lookback window, producing the normalized sequence $\bar{\mathbf{X}}_{1:L}$. The VARNN encoder and horizon decoder operate in the normalized space, and the stored statistics are used to transform the predicted horizon $\hat{\bar{\mathbf{Y}}}_{L+1:L+H}$ back to the original scale $\hat{\mathbf{Y}}_{L+1:L+H}$.

\subsubsection{Dual-State Recurrent Encoder}
\label{sec:varnn}

The encoder extracts temporal features from the past input window. We use VARNN~\cite{gharwi2025varnn} as the encoder that maintains two complementary recurrent states: a conventional hidden state and a residual-memory state. The hidden state captures the main temporal evolution of the sequence, while the residual-memory state tracks local deviations between one-step base predictions and observed values.

\paragraph{VARNN encoder.}
The normalized sequence $\bar{\mathbf{X}}_{1:L}$ is processed sequentially by the VARNN encoder. For each variable $i$ and time step $t$, the encoder maintains a hidden state $\mathbf{h}^{(i)}_t \in \mathbb{R}^{K}$ and a residual-memory state $\mathbf{e}^{(i)}_t \in \mathbb{R}^{J}$. The hidden state is updated using the current observation together with the previous hidden and residual-memory states:
\begin{equation} \mathbf{h}_t^{(i)} = \phi_h\left( \mathbf{W}_h \left[ \bar{x}_t^{(i)}; \mathbf{h}_{t-1}^{(i)}; \mathbf{e}_{t-1}^{(i)} \right] + \mathbf{b}_h \right), \end{equation}

where $\phi_h$ is a nonlinear activation function. The updated hidden state produces a one-step base prediction:
\begin{equation} \hat{\bar{x}}_{t+1}^{\mathrm{base},(i)} = \mathbf{w}_o^{\top}\mathbf{h}_t^{(i)} + b_o. \end{equation}
The corresponding local prediction residual is then computed as
\begin{equation} r_t^{(i)} = \bar{x}_{t+1}^{(i)} - \hat{\bar{x}}_{t+1}^{\mathrm{base},(i)}. \end{equation}
VARNN explicitly transforms this residual, together with the previous residual-memory state, through a nonlinear recurrent update:
\begin{equation} \mathbf{e}_t^{(i)} = \phi_e\left( \mathbf{W}_e \left[ r_t^{(i)}; \mathbf{e}_{t-1}^{(i)} \right] + \mathbf{b}_e \right), \end{equation}

where $\phi_e$ is a nonlinear activation function. Thus, $\mathbf{h}_t^{(i)}$ captures the primary temporal dynamics of the input sequence, while $\mathbf{e}_t^{(i)}$ explicitly tracks structured deviations between one-step base predictions and observed values. Unlike conventional recurrent models, which may encode such deviations implicitly within a single latent state, VARNN preserves them as a distinct recurrent signal that influences the subsequent hidden-state transition.

\paragraph{Encoder Temporal representation}
After processing the input window $\bar{\mathbf{X}}_{1:L}$, the encoder produces two temporal state trajectories \ignore{sequences}for each variable:

\begin{equation}
    \mathbf{H}^{(i)}
    =
    [\mathbf{h}_1^{(i)}, \mathbf{h}_2^{(i)}, \ldots, \mathbf{h}_{L-1}^{(i)}], \in  \mathbf{R}^{L-1 \times K}
\end{equation}
\begin{equation}
    \mathbf{E}^{(i)}
    =
    [\mathbf{e}_1^{(i)}, \mathbf{e}_2^{(i)}, \ldots, \mathbf{e}_{L-1}^{(i)}], \in \mathbf{R}^{ L-1 \times J}
\end{equation}

The hidden-state trajectory $\mathbf{H}^{(i)}$ captures the primary temporal dynamics of the input sequence, including trend, seasonality, level changes, and recurring temporal patterns. In contrast, the residual-memory trajectory captures structured local prediction deviations that may reflect drift, bias, or other non-stationary changes. These complementary trajectories are subsequently provided to the horizon forecasting decoder.

\subsection{Horizon Forecasting Decoder}


The horizon forecasting decoder maps the encoder state sequences to the future prediction horizon. Given the hidden-state sequence $\mathbf{H}^{(i)} \in \mathbb{R}^{(L-1) \times k}$ and the residual-memory sequence $\mathbf{E}^{(i)} \in \mathbb{R}^{(L-1) \times J}$ for variable $i$, the decoder processes the two sequences through separate temporal summarization branches. This design allows the model to preserve the distinct roles of the hidden and residual-memory states before combining them for long-horizon prediction $\hat{\mathbf{Y}}_{L+1:L+H}$.

\paragraph{Chunk Summarization Layer.}

The chunk summarization layer compresses the encoder state sequences into local temporal summaries before horizon projection. This reduces the parameter growth of direct flattening while preserving local patterns in the hidden-state and residual-memory sequences. In the main model, the hidden-state sequence $\mathbf{H}^{(i)}$ and the residual-memory sequence $\mathbf{E}^{(i)}$ are processed by separate chunk branches, allowing the decoder to preserve their distinct roles. Given a chunk window size $w$ and stride $s$, the starting index of the $m$-th chunk is defined as
$
    a_m = 1 + (m-1)s,  
$
    for
$
    m = 1,2,\ldots,M,
$
where the total number of chunks is
$
    M =
    \left\lfloor
    \frac{(L-1)-w}{s}
    \right\rfloor
    + 1.
$
For the hidden-state sequence, the $m$-th chunk is
$
    \mathbf{C}_{h,m}^{(i)}
    =
    \mathbf{H}_{a_m:a_m+w-1}^{(i)},
$
and for the residual-memory sequence, the corresponding chunk is
$
    \mathbf{C}_{e,m}^{(i)}
    =
    \mathbf{E}_{a_m:a_m+w-1}^{(i)}.
$
Each chunk is flattened and projected into a fixed-dimensional embedding:
\begin{equation}
    \mathbf{q}_{h,m}^{(i)}
    =
    \phi_{c,h}
    \left(
    \mathbf{W}_{c,h}
    \mathrm{vec}(\mathbf{C}_{h,m}^{(i)})
    +
    \mathbf{b}_{c,h}
    \right), \in \mathbf{R}^{T }
\end{equation}
\begin{equation}
    \mathbf{q}_{e,m}^{(i)}
    =
    \phi_{c,e}
    \left(
    \mathbf{W}_{c,e}
    \mathrm{vec}(\mathbf{C}_{e,m}^{(i)})
    +
    \mathbf{b}_{c,e}
    \right), \in \mathbf{R}^{I}
\end{equation}

The resulting chunk embeddings are collected as
$
    \mathbf{Q}_{h}^{(i)}
    =
    [\mathbf{q}_{h,1}^{(i)}, \ldots, \mathbf{q}_{h,M}^{(i)}], 
    \mathbf{Q}_{e}^{(i)}
    =
    [\mathbf{q}_{e,1}^{(i)}, \ldots, \mathbf{q}_{e,M}^{(i)}].
$
These summarized hidden and residual-memory representations are then passed to the forecasting head.

\paragraph{Forecasting Head }
The forecasting head maps the summarized hidden and residual-memory representations to the target prediction horizon. The chunk embeddings from the hidden and residual branches are first concatenated  $ \mathbf{Q}^{(i)} =  [\mathbf{Q}_{h}^{(i)}, \mathbf{Q}_{e}^{(i)}]$ and flattened , then passed through a linear projection to generate the normalized forecast for variable $i$:
\begin{equation}
    \hat{\bar{\mathbf{Y}}}_{L+1:L+H}^{(i)}
    =
    \mathbf{W}_{o}
    \mathrm{vec}(\mathbf{Q}^{(i)})
    +
    \mathbf{b}_{o}.
\end{equation}
Finally, The resulting output is later transformed back to the original scale using the inverse instance normalization step:  $\hat{\mathbf{Y}}_{L+1:L+H}^{(i)} \in \mathbb{R}^{1 \times  H}$.

\subsubsection{Training Objective and Optimization}
\label{sec:optimization}
The proposed model is optimized using a two-stage training procedure. In the first stage, the VARNN encoder is pretrained using one-step base prediction over the lookback length $L$. Given the one-step ahead base $\hat{\bar{\mathbf{x}}}_{t+1}$ and ground truth $\bar{\mathbf{x}}_{t+1}$, the encoder parameters are optimized using the mean squared error for each variable and averaged over all $D$ variables:

\begin{equation}
\mathcal{L}_{\mathrm{enc}}
=
\frac{1}{D}
\sum_{i=1}^{D}
\left\|
\hat{\bar{\mathbf{x}}}^{base,(i)}_{2:L}
-
\bar{\mathbf{x}}^{(i)}_{2:L}
\right\|_2^2.
\end{equation}

After pretraining, the encoder is frozen and a horizon decoder is trained to predict the future horizon sequence  $\hat{\mathbf{Y}}_{L+1:L+H}$. The forecasting objective is

\begin{equation}
\mathcal{L}_{\mathrm{forecast}}
=
\frac{1}{D}
\sum_{i=1}^{D}
\left\|
\hat{\mathbf{x}}^{(i)}_{L+1:L+H}
-
\mathbf{x}^{(i)}_{L+1:L+H}
\right\|_2^2.
\end{equation}

The two-stage strategy separates representation learning from horizon forecasting, provides an explicit training signal for the residual-memory state, and enables a pretrained encoder to be reused across multiple forecasting horizons for the same dataset and lookback length.

\subsection{Computational Complexity}
\label{sec:complexity}

Let $L$ be the lookback length, $D$ the number of variables, $H$ the forecasting horizon, and $d=d_h+d_e$ the combined encoder-state dimension. Given chunk window $w$ and stride $s$, the number of chunks is
$
    M = \left\lfloor \frac{L-w}{s} \right\rfloor + 1 .
$
Since $w$, $d_h$, and $d_e$ are fixed model hyperparameters, the dominant forward-pass complexity of VARNN can be summarized as
\begin{equation}
    \mathcal{O}\left(
    DLd^2
    +
    DMw(d_h^2+d_e^2)
    +
    DMdH
    \right).
\end{equation}
The first term corresponds to the recurrent encoder over the lookback window, while the second term corresponds to the chunk summarization layers and the last to the horizon projection head. Thus, VARNN scales linearly with the lookback length $L$ and linearly with the forecasting horizon $H$. Compared with Transformer-based forecasters, VARNN differs mainly in the representation encoder. Standard temporal self-attention has complexity $\mathcal{O}(L^2d)$ per layer, PatchTST reduces this to $\mathcal{O}(P^2d)$ using $P$ patches, and iTransformer applies attention over variables with complexity $\mathcal{O}(D^2d)$. Similar to these direct forecasting models, VARNN also has a horizon-dependent projection head. However, VARNN replaces attention-based representation learning with compact recurrent updates and local chunk summarization, yielding linear scaling with the lookback length $L$.

\section{Experiment Settings}
\label{sec:exp}
This section evaluates the proposed StateFlow, VARNN-based long-horizon forecasting framework, on standard multivariate time-series forecasting benchmarks. The experiments are designed to answer three main questions: (i) whether the proposed model is competitive with established long-horizon forecasting baselines, (ii) whether the hidden and residual-memory encoder sequences provide useful forecasting representations, and (iii) whether the chunk-based decoder improves the accuracy--complexity tradeoff compared with direct flattening.

\paragraph{Datasets }
We thoroughly evaluate the proposed StateFlow model on seven standard long-horizon forecasting benchmarks: ETTh1, ETTh2, ETTm1, ETTm2, Weather, ECL, and Traffic. The ETT datasets provide electricity transformer temperature measurements at hourly and minute-level resolutions \cite{zhou2021informer}. Weather consists of multivariate meteorological observations, ECL contains electricity consumption records from multiple clients, and Traffic includes road occupancy measurements collected from freeway sensors. These datasets are widely adopted in LTSF evaluation protocols\cite{wu2021autoformer, nie2023patchtst, liu2024itransformer}. \ignore{We set the lookback length to $L=96$ and evaluate horizons $H \in \{96,192,336,720\}$ for all datasets. }

\paragraph{Baselines }
We compare StateFlow with ten state-of-the-art deep forecasting models covering three major LTSF model families. The Transformer-based baselines include iTransformer, PatchTST, Crossformer, and Autoformer \cite{liu2024itransformer, nie2023patchtst, wu2021autoformer}. The convolutional baselines include TimesNet and SCINet \cite{wu2023timesnet}. The linear baselines include DLinear \cite{zeng2023dlinear} and RLinear \cite{li2023rlinear}. This set of baselines provides a broad comparison against attention-based, patch-based, convolutional, and lightweight linear forecasting architectures.

\paragraph{Experimental Setup }
\label{app: hyperparameters}
We use the standard LTSF setting with lookback length $L=96$ and prediction horizons $H \in \{96,192,336,720\}$. VARNN is implemented in a channel-independent manner, where each variable is modeled as a univariate sequence with shared parameters across variables. The experiment is implemented in PyTorch, with random seed 2026. The main model uses hidden dimension $k=32$, residual-memory dimension $J=16$, chunk summarizer dimensions $T=32, I=16$, chunk window size $w=5$, and stride $s=2$. We use ReLU activation for the hidden state and chunk summarizer, and tanh activation for the residual-memory state. Instance normalization is applied before the encoder, followed by inverse normalization after prediction. The model is trained in two stages. In stage 1, the encoder is pretrained with one-step base prediction loss. In Stage 2, the pretrained encoder is frozen, and the horizon forecasting decoder is trained using multi-step forecasting loss. We report MSE and MAE on the test set.

\section{Results}
\label{sec:results}

\begin{table*}[h]
\centering
\caption{Benchmark results (MSE/MAE) across datasets and horizons.}
\label{tab:main_results}
\scriptsize
\setlength{\tabcolsep}{3.0pt}
\renewcommand{\arraystretch}{1.15}
\resizebox{\textwidth}{!}{%
\begin{tabular}{|l|l||cc|cc|cc|cc|cc|cc|cc|cc|cc|cc|cc|}
\hline
\multirow{2}{*}{Dataset} & \multirow{2}{*}{H} &
\multicolumn{2}{c|}{StateFlow} &
\multicolumn{2}{c|}{iTrans.} &
\multicolumn{2}{c|}{RLinear} &
\multicolumn{2}{c|}{PatchTST} &
\multicolumn{2}{c|}{Cross.} &
\multicolumn{2}{c|}{TiDE} &
\multicolumn{2}{c|}{TimesNet} &
\multicolumn{2}{c|}{DLinear} &
\multicolumn{2}{c|}{SCINet} &
\multicolumn{2}{c|}{Stationary} &
\multicolumn{2}{c|}{Auto.} \\
\cline{3-4}\cline{5-6}\cline{7-8}\cline{9-10}\cline{11-12}\cline{13-14}\cline{15-16}\cline{17-18}\cline{19-20}\cline{21-22}\cline{23-24}
& &
MSE & MAE & MSE & MAE & MSE & MAE & MSE & MAE & MSE & MAE & MSE & MAE & MSE & MAE & MSE & MAE & MSE & MAE & MSE & MAE & MSE & MAE  \\
\hline\hline

\multirow{5}{*}{ECL}
& 96 & \underline{0.167} & \underline{0.257} & \textbf{0.148} & \textbf{0.240} & 0.201 & 0.281 & 0.195 & 0.285 & 0.219 & 0.314 & 0.237 & 0.329 & 0.168 & 0.272 & 0.197 & 0.282 & 0.247 & 0.345 & 0.169 & 0.273 & 0.201 & 0.317 \\
& 192 & \underline{0.177} & \underline{0.265} & \textbf{0.162} & \textbf{0.253} & 0.201 & 0.283 & 0.199 & 0.289 & 0.231 & 0.322 & 0.236 & 0.330 & 0.184 & 0.289 & 0.196 & 0.285 & 0.257 & 0.355 & 0.182 & 0.286 & 0.222 & 0.334 \\
& 336 & \underline{0.192} & \underline{0.280} & \textbf{0.178} & \textbf{0.269} & 0.215 & 0.298 & 0.215 & 0.305 & 0.246 & 0.337 & 0.249 & 0.344 & 0.198 & 0.300 & 0.209 & 0.301 & 0.269 & 0.369 & 0.200 & 0.304 & 0.231 & 0.338 \\
& 720 & 0.235 & \textbf{0.317} & 0.225 & \textbf{0.317} & 0.257 & 0.331 & 0.256 & 0.337 & 0.280 & 0.363 & 0.284 & 0.373 & \textbf{0.220} & \underline{0.320} & 0.245 & 0.333 & 0.299 & 0.390 & \underline{0.222} & 0.321 & 0.254 & 0.361 \\
& Avg & 0.193 & \underline{0.280} & \textbf{0.178} & \textbf{0.270} & 0.218 & 0.298 & 0.216 & 0.304 & 0.244 & 0.334 & 0.252 & 0.344 & \underline{0.193} & 0.295 & 0.212 & 0.300 & 0.268 & 0.365 & 0.193 & 0.296 & 0.227 & 0.338 \\
\hline
\multirow{5}{*}{Weather}
& 96 & \underline{0.163} & \textbf{0.212} & 0.174 & \underline{0.214} & 0.192 & 0.232 & 0.177 & 0.218 & \textbf{0.158} & 0.230 & 0.202 & 0.261 & 0.172 & 0.220 & 0.196 & 0.255 & 0.221 & 0.306 & 0.173 & 0.223 & 0.266 & 0.336 \\
& 192 & \underline{0.208} & \textbf{0.252} & 0.221 & \underline{0.254} & 0.240 & 0.271 & 0.225 & 0.259 & \textbf{0.206} & 0.277 & 0.242 & 0.298 & 0.219 & 0.261 & 0.237 & 0.296 & 0.261 & 0.340 & 0.245 & 0.285 & 0.307 & 0.367 \\
& 336 & \textbf{0.268} & \textbf{0.295} & 0.278 & \underline{0.296} & 0.292 & 0.307 & 0.278 & 0.297 & \underline{0.272} & 0.335 & 0.287 & 0.335 & 0.280 & 0.306 & 0.283 & 0.335 & 0.309 & 0.378 & 0.321 & 0.338 & 0.359 & 0.395 \\
& 720 & \textbf{0.341} & \textbf{0.343} & 0.358 & 0.349 & 0.364 & 0.353 & 0.354 & \underline{0.348} & 0.398 & 0.418 & 0.351 & 0.386 & 0.365 & 0.359 & \underline{0.345} & 0.381 & 0.377 & 0.427 & 0.414 & 0.410 & 0.419 & 0.428 \\
& Avg & \textbf{0.245} & \textbf{0.275} & \underline{0.258} & \underline{0.278} & 0.272 & 0.291 & 0.259 & 0.280 & 0.259 & 0.315 & 0.270 & 0.320 & 0.259 & 0.286 & 0.265 & 0.317 & 0.292 & 0.363 & 0.288 & 0.314 & 0.338 & 0.382 \\
\hline
\multirow{5}{*}{ETTh1}
& 96 & \textbf{0.364} & \textbf{0.394} & 0.386 & 0.405 & 0.386 & \underline{0.395} & 0.414 & 0.419 & 0.423 & 0.448 & 0.479 & 0.464 & \underline{0.384} & 0.402 & 0.386 & 0.400 & 0.654 & 0.599 & 0.513 & 0.491 & 0.449 & 0.459 \\
& 192 & \textbf{0.415} & \textbf{0.421} & 0.441 & 0.436 & 0.437 & \underline{0.424} & 0.460 & 0.445 & 0.471 & 0.474 & 0.525 & 0.492 & \underline{0.436} & 0.429 & 0.437 & 0.432 & 0.719 & 0.631 & 0.534 & 0.504 & 0.500 & 0.482 \\
& 336 & \textbf{0.452} & \textbf{0.440} & \underline{0.468} & 0.458 & 0.479 & \underline{0.446} & 0.501 & 0.466 & 0.570 & 0.546 & 0.565 & 0.515 & 0.491 & 0.469 & 0.481 & 0.459 & 0.778 & 0.659 & 0.588 & 0.535 & 0.521 & 0.496 \\
& 720 & \textbf{0.474} & \textbf{0.463} & 0.503 & 0.491 & \underline{0.481} & \underline{0.470} & 0.500 & 0.488 & 0.653 & 0.621 & 0.594 & 0.558 & 0.521 & 0.500 & 0.519 & 0.516 & 0.836 & 0.699 & 0.643 & 0.616 & 0.514 & 0.512 \\
& Avg & \textbf{0.426} & \textbf{0.429} & 0.450 & 0.448 & \underline{0.446} & \underline{0.434} & 0.469 & 0.455 & 0.529 & 0.522 & 0.541 & 0.507 & 0.458 & 0.450 & 0.456 & 0.452 & 0.747 & 0.647 & 0.570 & 0.536 & 0.496 & 0.487 \\
\hline
\multirow{5}{*}{ETTh2}
& 96 & \underline{0.289} & \underline{0.344} & 0.297 & 0.349 & \textbf{0.288} & \textbf{0.338} & 0.302 & 0.348 & 0.745 & 0.584 & 0.400 & 0.440 & 0.340 & 0.374 & 0.333 & 0.387 & 0.707 & 0.621 & 0.476 & 0.458 & 0.346 & 0.388 \\
& 192 & \textbf{0.371} & \underline{0.395} & 0.380 & 0.400 & \underline{0.374} & \textbf{0.390} & 0.388 & 0.400 & 0.877 & 0.656 & 0.528 & 0.509 & 0.402 & 0.414 & 0.477 & 0.476 & 0.860 & 0.689 & 0.512 & 0.493 & 0.456 & 0.452 \\
& 336 & \underline{0.419} & \underline{0.430} & 0.428 & 0.432 & \textbf{0.415} & \textbf{0.426} & 0.426 & 0.433 & 1.043 & 0.731 & 0.643 & 0.571 & 0.452 & 0.452 & 0.594 & 0.541 & 1.000 & 0.744 & 0.552 & 0.551 & 0.482 & 0.486 \\
& 720 & 0.432 & 0.448 & \underline{0.427} & \underline{0.445} & \textbf{0.420} & \textbf{0.440} & 0.431 & 0.446 & 1.104 & 0.763 & 0.874 & 0.679 & 0.462 & 0.468 & 0.831 & 0.657 & 1.249 & 0.838 & 0.562 & 0.560 & 0.515 & 0.511 \\
& Avg & \underline{0.378} & \underline{0.404} & 0.383 & 0.407 & \textbf{0.374} & \textbf{0.398} & 0.387 & 0.407 & 0.942 & 0.683 & 0.611 & 0.550 & 0.414 & 0.427 & 0.559 & 0.515 & 0.954 & 0.723 & 0.526 & 0.516 & 0.450 & 0.459 \\
\hline
\multirow{5}{*}{ETTm1}
& 96 & \textbf{0.316} & \textbf{0.359} & 0.334 & 0.368 & 0.355 & 0.376 & \underline{0.329} & \underline{0.367} & 0.404 & 0.426 & 0.364 & 0.387 & 0.338 & 0.375 & 0.345 & 0.372 & 0.418 & 0.438 & 0.386 & 0.398 & 0.505 & 0.475 \\
& 192 & \textbf{0.351} & \textbf{0.378} & 0.387 & 0.391 & 0.391 & 0.392 & \underline{0.367} & \underline{0.385} & 0.450 & 0.451 & 0.398 & 0.404 & 0.374 & 0.387 & 0.380 & 0.389 & 0.426 & 0.441 & 0.459 & 0.444 & 0.553 & 0.496 \\
& 336 & \textbf{0.382} & \textbf{0.400} & 0.426 & 0.420 & 0.424 & 0.415 & \underline{0.399} & \underline{0.410} & 0.532 & 0.515 & 0.428 & 0.425 & 0.410 & 0.411 & 0.413 & 0.413 & 0.445 & 0.459 & 0.495 & 0.464 & 0.621 & 0.537 \\
& 720 & \textbf{0.445} & \textbf{0.437} & 0.491 & 0.459 & 0.487 & 0.450 & \underline{0.454} & \underline{0.439} & 0.666 & 0.589 & 0.487 & 0.461 & 0.478 & 0.450 & 0.474 & 0.453 & 0.595 & 0.550 & 0.585 & 0.516 & 0.671 & 0.561 \\
& Avg & \textbf{0.373} & \textbf{0.394} & 0.409 & 0.410 & 0.414 & 0.408 & \underline{0.387} & \underline{0.400} & 0.513 & 0.495 & 0.419 & 0.419 & 0.400 & 0.406 & 0.403 & 0.407 & 0.471 & 0.472 & 0.481 & 0.456 & 0.588 & 0.517 \\
\hline
\multirow{5}{*}{ETTm2}
& 96 & \textbf{0.175} & \underline{0.262} & \underline{0.180} & 0.264 & 0.182 & 0.265 & \textbf{0.175} & \textbf{0.259} & 0.287 & 0.366 & 0.207 & 0.305 & 0.187 & 0.267 & 0.193 & 0.292 & 0.286 & 0.377 & 0.192 & 0.274 & 0.255 & 0.339 \\
& 192 & \textbf{0.241} & \underline{0.304} & 0.250 & 0.309 & \underline{0.246} & \underline{0.304} & \textbf{0.241} & \textbf{0.302} & 0.414 & 0.492 & 0.290 & 0.364 & 0.249 & 0.309 & 0.284 & 0.362 & 0.399 & 0.445 & 0.280 & 0.339 & 0.281 & 0.340 \\
& 336 & \textbf{0.300} & \textbf{0.342} & 0.311 & 0.348 & 0.307 & \textbf{0.342} & \underline{0.305} & \underline{0.343} & 0.597 & 0.542 & 0.377 & 0.422 & 0.321 & 0.351 & 0.369 & 0.427 & 0.637 & 0.591 & 0.334 & 0.361 & 0.339 & 0.372 \\
& 720 & \textbf{0.396} & 0.401 & 0.412 & 0.407 & 0.407 & \textbf{0.398} & \underline{0.402} & \underline{0.400} & 1.730 & 1.042 & 0.558 & 0.524 & 0.408 & 0.403 & 0.554 & 0.522 & 0.960 & 0.735 & 0.417 & 0.413 & 0.433 & 0.432 \\
& Avg & \textbf{0.278} & \underline{0.327} & 0.288 & 0.332 & 0.285 & \underline{0.327} & \underline{0.281} & \textbf{0.326} & 0.757 & 0.611 & 0.358 & 0.404 & 0.291 & 0.333 & 0.350 & 0.401 & 0.571 & 0.537 & 0.306 & 0.347 & 0.327 & 0.371 \\
\hline
\multirow{5}{*}{Traffic}
& 96 & 0.569 & 0.364 & \textbf{0.395} & \textbf{0.268} & 0.649 & 0.389 & \underline{0.462} & 0.295 & 0.522 & \underline{0.290} & 0.805 & 0.493 & 0.593 & 0.321 & 0.650 & 0.396 & 0.788 & 0.499 & 0.612 & 0.338 & 0.613 & 0.388 \\
& 192 & 0.540 & 0.345 & \textbf{0.417} & \textbf{0.276} & 0.601 & 0.366 & \underline{0.466} & 0.296 & 0.530 & \underline{0.293} & 0.756 & 0.474 & 0.617 & 0.336 & 0.598 & 0.370 & 0.789 & 0.505 & 0.613 & 0.340 & 0.616 & 0.382 \\
& 336 & 0.537 & 0.338 & \textbf{0.433} & \textbf{0.283} & 0.609 & 0.369 & \underline{0.482} & \underline{0.304} & 0.558 & 0.305 & 0.762 & 0.477 & 0.629 & 0.336 & 0.605 & 0.373 & 0.797 & 0.508 & 0.618 & 0.328 & 0.622 & 0.337 \\
& 720 & 0.541 & 0.346 & \textbf{0.467} & \textbf{0.302} & 0.647 & 0.387 & \underline{0.514} & \underline{0.322} & 0.589 & 0.328 & 0.719 & 0.449 & 0.640 & 0.350 & 0.645 & 0.394 & 0.841 & 0.523 & 0.653 & 0.355 & 0.660 & 0.408 \\
& Avg & 0.547 & 0.348 & \textbf{0.428} & \textbf{0.282} & 0.627 & 0.378 & \underline{0.481} & 0.304 & 0.550 & \underline{0.304} & 0.760 & 0.473 & 0.620 & 0.336 & 0.625 & 0.383 & 0.804 & 0.509 & 0.624 & 0.340 & 0.628 & 0.379 \\
\hline\hline
\multicolumn{2}{|l||}{1st Count} & 15 & 14 & 7 & 8 & 3 & 6 & 2 & 2 & 2 & 0 & 0 & 0 & 1 & 0 & 0 & 0 & 0 & 0 & 0 & 0 & 0 & 0 \\
\hline
\end{tabular}%
}
\vspace{1mm}
\textit{Note: The baseline results are taken from the standardized benchmark table reported by iTransformer \cite{liu2024itransformer}, and implemented under the same standard training protocols: input length $L=96$, prediction horizons $H \in \{96,192,336,720\}$, and MSE/MAE evaluation.} 
\end{table*}

\textbf{Overall performance } 
Table~\ref{tab:main_results} shows the multivariate long term forecasting results. A lower MSE or MAE indicates better forecasting performance. Overall, StateFlow achieves competitive performance compared with strong linear, convolutional, and Transformer-based models and obtains the largest number of first-place results, with 15 MSE wins and 14 MAE wins across 28 dataset-horizon settings. The results suggested that the proposed residual-aware recurrent encoder can provide effective long-horizon forecasting representations, despite using a structurally different design from patch-based and attention-based architectures. In particular, the hidden and residual-memory state sequences allow the decoder to exploit both primary temporal dynamics and structured prediction deviations when generating the future horizon. On high-dimensional datasets such as ECL and Traffic, iTransformer remains stronger in several settings. These high-dimensional datasets may benefit more from explicit cross-variate dependency modeling, whereas StateFlow adopts a channel-independent design and focuses primarily on temporal modeling within each variable. Overall, the results indicate that residual-memory recurrence is an effective and efficient inductive bias for long-horizon forecasting, while incorporating stronger cross-variate interactions may further improve performance on high-dimensional multivariate datasets.

\subsection{Ablations}
\label{sec:ablation}

\subsubsection{Effect of VARNN Encoder Representation}

Table~\ref{tab:encoder_representation_ablation} compares four encoder representations under the same chunk-decoder setting: the dual-state representation $(h,e)$, the hidden-state sequence $h$, the residual-memory sequence $e$, and the base-prediction with corresponding residual scalar representation $(\hat{\bar{\mathbf{x}}}^{base},r)$. Overall, the dual-state representation $(h,e)$ provides the most stable and effective forecasting representation, achieving the best results on most horizons, especially on ETTh1, ETTm1, and ETTm2. This shows that using both the residual-conditioned hidden state and the residual-memory state gives the decoder richer information for long-horizon prediction. The individual representations clarify the role of each state. The hidden sequence $h$ remains competitive, indicating that it captures strong recurrent dynamics conditioned on past residual-memory information. The residual-memory sequence $e$ performs particularly well on ETTh2 and is often second-best on ETTm2, suggesting that structured prediction deviations can provide useful forecasting signals. In contrast, $(\hat{\bar{\mathbf{x}}}^{base},r)$ is generally weaker than the learned state representations, showing that the nonlinear recurrent encoding of prediction errors is more effective than directly passing raw base predictions and residuals. These results support the main StateFlow design, whereas hidden and residual-memory states provide complementary information, and their dual-state representation improves robustness across datasets and horizons.

\paragraph{Dual-State Fusion Strategy} To examine how the dual-state representation should be fused, we compare separate and joint chunk summarization in Table~\ref{tab:dual_state_fusion_ablation}. The separate chunking $(h,e)$ summarizes the hidden trajectory $h$ and residual-memory trajectory $e$ using independent chunk summarizers before concatenation, while the joint concatenates $[h,e]$ before chunk summarization. Separate chunking achieves lower average MSE and MAE in most datasets and horizons, and uses fewer parameters. This suggests that preserving the distinction between hidden dynamics and residual-memory dynamics before temporal compression is beneficial. Full results are reported in Appendix~\ref{app:dual_state_fusion_full}.

\begin{table}[t]
\centering
\scriptsize
\caption{Ablation study of encoder representations under the fixed chunk-decoder setting. Lower values are better. The best result is shown in bold and the second-best result is underlined.}
\label{tab:encoder_representation_ablation}
\begin{tabular}{l|l|rr|rr|rr|rr}
\hline
Dataset & H & \multicolumn{2}{c|}{$(h,e)$} & \multicolumn{2}{c|}{h} & \multicolumn{2}{c|}{e} & \multicolumn{2}{c}{$(\hat{\bar{\mathbf{x}}}^{base},r)$} \\
\cline{3-10}
 &  & MSE & MAE & MSE & MAE & MSE & MAE & MSE & MAE \\
\hline
ETTh1 & 96 & \textbf{0.364} & \textbf{0.394} & \underline{0.367} & 0.395 & 0.375 & \underline{0.394} & 0.392 & 0.398 \\
 & 192 & \textbf{0.415} & \textbf{0.421} & \underline{0.418} & \underline{0.424} & 0.435 & 0.425 & 0.444 & 0.426 \\
 & 336 & \underline{0.452} & \textbf{0.440} & \textbf{0.445} & \underline{0.442} & 0.478 & 0.445 & 0.466 & 0.443 \\
 & 720 & \textbf{0.474} & \textbf{0.463} & 0.488 & 0.469 & 0.493 & 0.473 & \underline{0.482} & \underline{0.465} \\
\hline
ETTh2 & 96 & \textbf{0.289} & \textbf{0.344} & 0.299 & 0.348 & \underline{0.290} & \underline{0.345} & 0.304 & 0.352 \\
 & 192 & 0.371 & \underline{0.395} & 0.372 & 0.397 & \textbf{0.364} & \textbf{0.389} & \underline{0.371} & 0.395 \\
 & 336 & \textbf{0.419} & \textbf{0.430} & 0.431 & 0.436 & \underline{0.422} & \underline{0.433} & 0.432 & 0.438 \\
 & 720 & \underline{0.432} & \underline{0.448} & 0.437 & 0.451 & \textbf{0.430} & \textbf{0.445} & 0.444 & 0.453 \\
\hline
ETTm1 & 96 & \textbf{0.316} & \textbf{0.359} & \underline{0.323} & \underline{0.365} & 0.334 & 0.370 & 0.353 & 0.375 \\
 & 192 & \textbf{0.351} & \textbf{0.378} & \underline{0.353} & \underline{0.379} & 0.368 & 0.386 & 0.384 & 0.390 \\
 & 336 & \textbf{0.382} & \textbf{0.400} & \underline{0.385} & \underline{0.402} & 0.396 & 0.403 & 0.415 & 0.408 \\
 & 720 & \underline{0.445} & \underline{0.438} & \textbf{0.443} & \textbf{0.436} & 0.448 & 0.438 & 0.474 & 0.441 \\
\hline
ETTm2 & 96 & \textbf{0.175} & \textbf{0.262} & \underline{0.180} & 0.265 & 0.180 & 0.267 & 0.181 & \underline{0.264} \\
 & 192 & \textbf{0.241} & \textbf{0.304} & 0.248 & 0.312 & \underline{0.243} & \underline{0.307} & 0.246 & 0.310 \\
 & 336 & \textbf{0.300} & \textbf{0.342} & 0.307 & 0.349 & \underline{0.305} & 0.348 & 0.307 & \underline{0.345} \\
 & 720 & \textbf{0.396} & \textbf{0.401} & 0.407 & 0.405 & \underline{0.404} & \underline{0.402} & 0.405 & 0.403 \\
\hline
\end{tabular}
\end{table}

\begin{table}[t]
\centering
\scriptsize
\caption{Effect of dual-state fusion strategy under the chunk decoder. Lower values are better.}
\label{tab:dual_state_fusion_ablation}
\begin{tabular}{lcccc}
\toprule
Fusion strategy & Avg. MSE & Avg. MAE & MSE wins & Params \\
\midrule
Separate chunking $(h,e)$ & \textbf{0.364} & \textbf{0.389} & \textbf{12/16} & \textbf{750.6K} \\
Joint chunking $[h,e]$ & 0.366 & 0.390 & 4/16 & 755.7K \\
\bottomrule
\end{tabular}
\end{table}

\begin{figure*}[ht]
    \centering
    \includegraphics[width=0.85\textwidth]{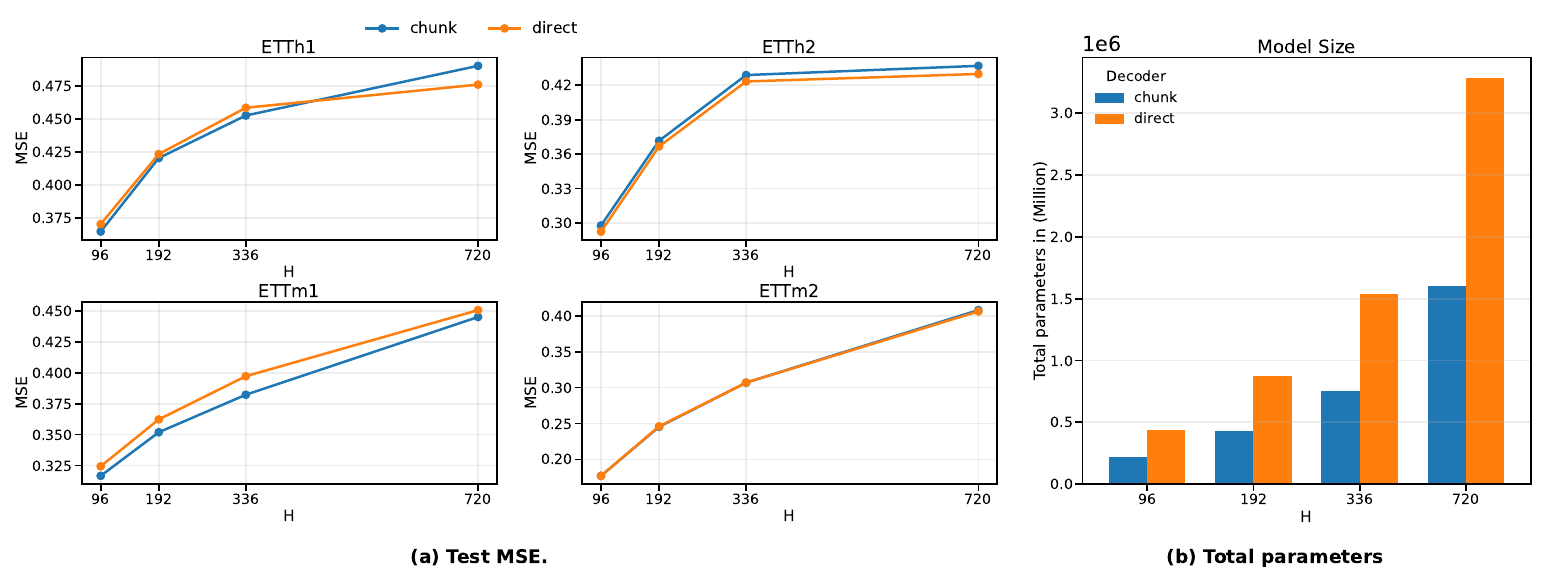}
    \caption{Effect of chunk versus direct decoding on the \textit{he} representation. The left panel reports test MSE across prediction horizons, while the right panel compares total parameters.}
    \label{fig:chunk_vs_direct_decoder}
\end{figure*}

\paragraph{Chunk vs Direct Decoder}
Figure~\ref{fig:chunk_vs_direct_decoder} compares the chunk decoder with direct flattening under the same hidden--residual representation. The chunk decoder achieves comparable or better MSE on most datasets, especially on ETTm1 and ETTh2, while direct decoding is only slightly better in some long-horizon ETTh1 cases. More importantly, the chunk decoder substantially reduces the model size by nearly half compared with direct decoding. These results show that the chunk summarization layer achieves similar or better accuracy with dramatically fewer parameters, so it is used as the default decoder in StateFlow.

\par\medskip

\subsection{Effect of Recurrent Encoder and Optimization Design}
\label{sec:Recurrent_Encoder_and_Optimization_Design}

\paragraph{Recurrent Encoder Choice}
We evaluate whether the proposed dual-state VARNN encoder provides advantages over standard recurrent encoders by comparing VARNN with RNN, GRU, and LSTM under the same recurrent encoder–decoder framework. All models use the same normalization, lookback length, chunk decoder, forecasting horizons. and two-stage optimization protocol. To control decoder capacity, we use a matched-representation setting: RNN, GRU, and LSTM use a 48-dimensional hidden state, while VARNN uses a 32-dimensional hidden state and a 16-dimensional residual-memory state, giving all models the same 48-dimensional decoder input. 

The results show that VARNN provides the best accuracy–efficiency tradeoff. Across the four ETT datasets and four horizons, VARNN achieves the lowest average error among recurrent encoders, with $0.364$ MSE, compared with LSTM $0.374$, RNN $0.376$, and GRU $0.377$. 
In terms of encoder complexity, VARNN uses the fewest encoder parameters: $1{,}922$ compared with $2{,}497$ for RNN, $7{,}393$ for GRU, and $9{,}841$ for LSTM. 
Thus, VARNN reduces encoder parameters by $23.03\%$ compared to RNN and
 $ 80.47\%$ compared with LSTM\ignore{while improving average MSE by approximately $3\%$}. These findings suggest that the residual-memory state improves recurrent forecasting by providing a more parameter-efficient temporal representation than standard gated recurrence.

\par\medskip

\paragraph{Optimization Strategy}
We further evaluate whether the proposed two-stage optimization contributes to the observed gains. In one-stage training, the encoder and horizon decoder are optimized jointly using the multi-step forecasting loss. In two-stage training, the encoder is first pretrained using one-step prediction and then frozen while the horizon decoder is trained. VARNN achieves the best average MSE under both settings, obtaining $0.371$ with one-stage training and $0.364$ with two-stage training strategy. The two-stage strategy also improves RNN ($0.378 \rightarrow 0.376$) and LSTM ($0.377 \rightarrow 0.374$), while GRU shows a small degradation ($0.376 \rightarrow 0.377$). The largest improvement is observed for VARNN, suggesting that one-step pretraining is particularly effective for learning residual-memory dynamics before long-horizon decoding. Moreover, since Stage~1 depends only on the dataset and lookback length, the pretrained encoder can be reused across multiple forecasting horizons, avoiding redundant encoder training and enabling horizon-specific decoders to be learned from a shared temporal representation. Full results are reported in Appendix~\ref{app:appendix_full_recurrent_ablation}, Table~\ref{tab:appendix_full_recurrent_ablation_table}.

\begin{figure}[t]
\centering

\begin{adjustbox}{valign=t,minipage=0.48\linewidth}
\centering

\captionof{table}{Two-stage recurrent encoder results under the matched-representation setting. Lower MSE is better.}
\label{tab:two_stage_recurrent_results}
\scriptsize

\begin{tabular}{lccccc}
\toprule
Model & ETTh1 & ETTh2 & ETTm1 & ETTm2 & Avg \\
\midrule
VARNN & \textbf{0.427} & \textbf{0.378} & \textbf{0.374} & \underline{0.279} & \textbf{0.364} \\
LSTM  & \underline{0.444} & \underline{0.385} & 0.384 & 0.283 & \underline{0.374} \\
RNN   & 0.446 & 0.391 & 0.386 & 0.281 & 0.376 \\
GRU   & 0.458 & 0.396 & \underline{0.378} & \textbf{0.278} & 0.377 \\
\bottomrule
\end{tabular}

\vspace{0.55em}

\captionof{table}{One-stage versus two-stage optimization performance under the matched-representation setting. Lower Avg MSE is better.}
\label{tab:one_vs_two_stage_recurrent}
\scriptsize
\begin{tabular}{llll}
\toprule
Model & One-stage & Two-stage & Two-stage gain \\
\midrule
VARNN & \textbf{0.371} & \textbf{0.364} & \textbf{+1.88\% $\uparrow$} \\
LSTM & 0.377 & \underline{0.374} & \underline{+0.97\% $\uparrow$} \\
RNN & 0.378 & 0.376 & +0.43\% $\uparrow$ \\
GRU & \underline{0.376} & 0.377 & -0.38\% $\downarrow$ \\
\bottomrule

\end{tabular}

\end{adjustbox}
\hfill
\begin{adjustbox}{valign=t,minipage=0.48\linewidth}
\centering
\includegraphics[width=\linewidth]{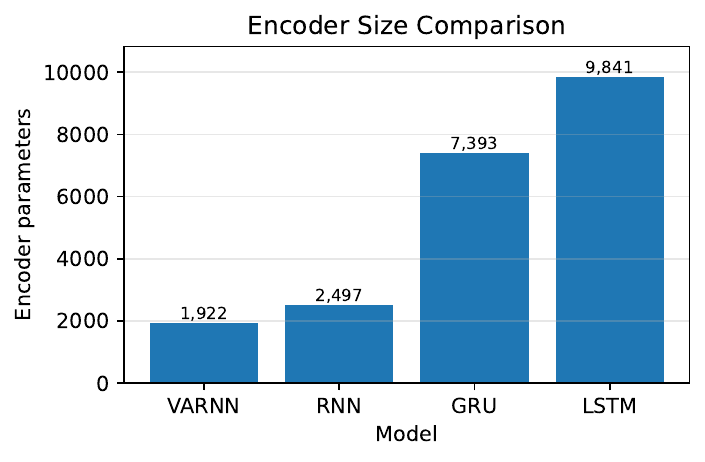}
\captionof{figure}{Encoder-only parameter comparison under the matched-representation setting. VARNN uses the smallest encoder.}
\label{fig:encoder_size_comparison}
\end{adjustbox}

\end{figure}
\par\medskip
\par\medskip

\section{Ethics Statement}
This work develops supervised learning methods for long-horizon time-series forecasting. Ethical risks are low. However, forecasting models can affect downstream decisions in domains such as energy, transportation, finance, and healthcare. When applied to sensitive domains, ensure appropriate consent procedures, privacy safeguards, fairness audits, and domain-specific human oversight.

\section{Conclusion and Future Work}
\label{sec:conclusion} 
This paper proposes StateFlow, a residual-aware recurrent framework for long-horizon time-series forecasting. The model extends VARNN by using hidden-state and residual-memory sequences as forecasting representations, enabling the decoder to leverage both primary temporal dynamics and structured residual patterns to predict the future horizon. Experiments on standard LTSF benchmarks show that StateFlow is competitive with strong linear, convolutional, and Transformer-based baselines. StateFlow provides an alternative to Transformer-based forecasting models by avoiding quadratic temporal self-attention. Its recurrent encoder scales linearly with the lookback length, while the chunk-based decoder reduces the parameter growth associated with directly flattening the full encoder state sequence. The results suggest that residual-aware recurrent modeling provides a promising direction for long-horizon forecasting. Future work will investigate cross-variate dependency modeling and more time-series analysis tasks.

\clearpage
\bibliographystyle{iclr2026_conference}
\bibliography{references.bib}
\clearpage
\appendix

\section{MORE RESULTS ON ABLATION STUDY}

\subsection{Full Dual-State Fusion Strategy Results} \label{app:dual_state_fusion_full} Table~\ref{tab:dual_state_fusion_full} reports the full dataset-horizon comparison between separate and joint dual-state fusion under the chunk decoder. The separate variant applies independent chunk summarizers to $h$ and $e$ before concatenation, while the joint variant concatenates $[h,e]$ before chunk summarization. Overall, separate chunking provides lower error in most settings while also using fewer parameters.

\begin{table}[h]
\centering
\scriptsize
\caption{Effect of dual-state fusion strategy under the fixed chunk-decoder setting. Lower MSE/MAE and fewer parameters are better. Extra Params reports the additional parameters used by Joint HE relative to Sep. HE.}
\label{tab:dual_state_fusion_full}
\begin{tabular}{l|l|rrr|rrr|c}
\hline
Dataset & H & \multicolumn{3}{c|}{Sep. HE $(h,e)$} & \multicolumn{3}{c|}{Joint HE $[h,e]$} & \multicolumn{1}{c}{Extra Params} \\
\cline{3-9}
 &  & MSE & MAE & Params & MSE & MAE & Params & Joint--Sep. \\
\hline
ETTh1 & 96 & \textbf{0.364} & \textbf{0.394} & \textbf{220.4K} & 0.368 & 0.395 & 225.6K & +5.1K \\
 & 192 & \textbf{0.415} & \textbf{0.421} & \textbf{432.5K} & 0.418 & 0.425 & 437.6K & +5.1K \\
 & 336 & \textbf{0.452} & \textbf{0.440} & \textbf{750.6K} & 0.465 & 0.445 & 755.7K & +5.1K \\
 & 720 & 0.474 & 0.463 & \textbf{1598.9K} & \textbf{0.469} & \textbf{0.461} & 1604.0K & +5.1K \\
\hline
ETTh2 & 96 & \textbf{0.289} & \textbf{0.344} & \textbf{220.4K} & 0.291 & 0.345 & 225.6K & +5.1K \\
 & 192 & \textbf{0.371} & \textbf{0.395} & \textbf{432.5K} & 0.382 & 0.401 & 437.6K & +5.1K \\
 & 336 & 0.419 & 0.430 & \textbf{750.6K} & \textbf{0.415} & \textbf{0.428} & 755.7K & +5.1K \\
 & 720 & \textbf{0.432} & 0.448 & \textbf{1598.9K} & 0.433 & \textbf{0.447} & 1604.0K & +5.1K \\
\hline
ETTm1 & 96 & 0.316 & 0.359 & \textbf{220.4K} & \textbf{0.314} & \textbf{0.357} & 225.6K & +5.1K \\
 & 192 & 0.351 & \textbf{0.378} & \textbf{432.5K} & \textbf{0.351} & 0.379 & 437.6K & +5.1K \\
 & 336 & \textbf{0.382} & \textbf{0.400} & \textbf{750.6K} & 0.382 & 0.402 & 755.7K & +5.1K \\
 & 720 & \textbf{0.445} & \textbf{0.438} & \textbf{1598.9K} & 0.445 & 0.439 & 1604.0K & +5.1K \\
\hline
ETTm2 & 96 & \textbf{0.175} & 0.262 & \textbf{220.4K} & 0.176 & \textbf{0.261} & 225.6K & +5.1K \\
 & 192 & \textbf{0.241} & \textbf{0.304} & \textbf{432.5K} & 0.241 & 0.305 & 437.6K & +5.1K \\
 & 336 & \textbf{0.300} & \textbf{0.342} & \textbf{750.6K} & 0.302 & 0.343 & 755.7K & +5.1K \\
 & 720 & \textbf{0.396} & \textbf{0.401} & \textbf{1598.9K} & 0.401 & 0.403 & 1604.0K & +5.1K \\
\hline
Overall & Avg & \textbf{0.364} & \textbf{0.389} & \textbf{750.6K} & 0.366 & 0.390 & 755.7K & +5.1K \\
Overall & Wins & 12/16 & 11/16 & -- & 4/16 & 5/16 & -- & -- \\
\hline
\end{tabular}
\end{table}

\clearpage

\subsection {Recurrent Encoder Choice and Optimization Strategies Results} 
\label{app:appendix_full_recurrent_ablation}
This appendix provides the full recurrent encoder and optimization ablation under the matched-representation setting. RNN, GRU, and LSTM use a 48-dimensional hidden state, while VARNN uses a 32-dimensional hidden state and a 16-dimensional residual-memory state, giving all models the same 48-dimensional decoder representation. All models use the same normalization, chunk decoder, lookback length, forecasting horizons, and evaluation protocol.

Table~\ref{tab:appendix_full_recurrent_ablation_table} reports MSE for each dataset and forecasting horizon under both one-stage and two-stage optimization. In one-stage training, the encoder and decoder are optimized jointly using the horizon forecasting loss. In two-stage training, the encoder is first pretrained using one-step prediction and then frozen while the horizon decoder is trained. The results show that VARNN achieves the best overall average under both optimization settings. The two-stage setting provides the strongest VARNN performance, supporting the role of one-step encoder pretraining in learning useful residual-memory dynamics before long-horizon decoding.

\begin{table*}[h]
\centering
\scriptsize
\setlength{\tabcolsep}{5.5pt}
\renewcommand{\arraystretch}{1.15}
\caption{Full recurrent encoder and optimization ablation under the matched-representation setting. Results are reported as MSE. Lower values are better. The best result is shown in bold and the second-best result is underlined.}
\label{tab:appendix_full_recurrent_ablation_table}
\begin{tabular}{l!{\vrule width 0.8pt}c!{\vrule width 0.8pt}cccc!{\vrule width 1.2pt}cccc}
\hline
\multicolumn{1}{c!{\vrule width 0.8pt}}{} & \multicolumn{1}{c!{\vrule width 0.8pt}}{} & \multicolumn{4}{c!{\vrule width 1.2pt}}{One-stage} & \multicolumn{4}{c}{Two-stage} \\
Dataset & H & VARNN & LSTM & RNN & GRU & VARNN & LSTM & RNN & GRU \\
\hline
ETTh1 & 96 & \underline{0.378} & 0.392 & \textbf{0.374} & 0.385 & \textbf{0.364} & \underline{0.380} & 0.382 & 0.384 \\
 & 192 & \underline{0.429} & 0.444 & \textbf{0.423} & 0.441 & \textbf{0.415} & 0.433 & \underline{0.421} & 0.437 \\
 & 336 & 0.484 & \underline{0.478} & \textbf{0.471} & 0.500 & \textbf{0.452} & \underline{0.465} & 0.470 & 0.494 \\
 & 720 & \textbf{0.469} & 0.503 & 0.526 & \underline{0.489} & \textbf{0.474} & \underline{0.497} & 0.508 & 0.513 \\
 & Avg & \textbf{0.440} & 0.454 & \underline{0.448} & 0.454 & \textbf{0.426} & \underline{0.444} & 0.445 & 0.457 \\
\hline
ETTh2 & 96 & \textbf{0.291} & \underline{0.294} & 0.305 & 0.300 & \textbf{0.289} & 0.300 & \underline{0.297} & 0.303 \\
 & 192 & 0.374 & \underline{0.371} & 0.372 & \textbf{0.369} & \textbf{0.371} & 0.374 & 0.375 & \underline{0.373} \\
 & 336 & 0.429 & 0.427 & \textbf{0.416} & \underline{0.420} & \underline{0.419} & \textbf{0.417} & 0.446 & 0.482 \\
 & 720 & \textbf{0.437} & 0.454 & 0.464 & \underline{0.452} & \underline{0.432} & 0.445 & 0.444 & \textbf{0.424} \\
 & Avg & \textbf{0.383} & 0.386 & 0.389 & \underline{0.385} & \textbf{0.378} & \underline{0.384} & 0.391 & 0.396 \\
\hline
ETTm1 & 96 & \textbf{0.317} & 0.331 & 0.344 & \underline{0.327} & \textbf{0.316} & 0.327 & 0.335 & \underline{0.323} \\
 & 192 & \underline{0.361} & 0.365 & \textbf{0.360} & 0.368 & \textbf{0.351} & 0.367 & 0.364 & \underline{0.359} \\
 & 336 & \textbf{0.386} & 0.395 & 0.390 & \underline{0.388} & \textbf{0.382} & 0.389 & 0.393 & \underline{0.384} \\
 & 720 & \textbf{0.441} & 0.467 & 0.452 & \underline{0.449} & \textbf{0.445} & 0.451 & 0.450 & \underline{0.446} \\
 & Avg & \textbf{0.376} & 0.389 & 0.387 & \underline{0.383} & \textbf{0.373} & 0.384 & 0.385 & \underline{0.378} \\
\hline
ETTm2 & 96 & 0.181 & \textbf{0.176} & 0.182 & \underline{0.179} & \textbf{0.175} & 0.179 & \underline{0.177} & \textbf{0.175} \\
 & 192 & 0.243 & \textbf{0.239} & 0.249 & \underline{0.242} & \underline{0.241} & 0.249 & 0.243 & \textbf{0.240} \\
 & 336 & 0.305 & \textbf{0.301} & 0.304 & \underline{0.301} & \textbf{0.300} & 0.302 & 0.305 & \textbf{0.300} \\
 & 720 & 0.409 & \textbf{0.395} & 0.403 & \underline{0.399} & \underline{0.396} & 0.397 & 0.399 & \textbf{0.395} \\
 & Avg & 0.284 & \textbf{0.278} & 0.284 & \underline{0.280} & \underline{0.278} & 0.282 & 0.281 & \textbf{0.277} \\
\hline
Overall & Avg & \textbf{0.371} & 0.377 & 0.377 & \underline{0.376} & \textbf{0.364} & \underline{0.373} & 0.375 & 0.377 \\
Overall & Wins & \textbf{6} & 4 & \underline{5} & 1 & \textbf{12} & 1 & 0 & \underline{5} \\
\hline
\end{tabular}
\end{table*}

\end{document}